\documentclass[twocolumn]{rps-esrel2026}

\def\papername{\jobname}

\usepackage{natbib}
\usepackage{algorithm}
\usepackage{algpseudocode}
\usepackage{graphicx}
\usepackage{subcaption}
\usepackage{tabularx}

\begin{document}

\markboth{Nehal Afifi and Abdelmonem Elhendawi}{Instructions for Preparing Paper for ESREL2026}

\twocolumn[

\title{Anomaly Detection for Electro-Hydrostatic Actuators using LSTM Autoencoder.}

\author{Nehal Afifi}
\address{IPEK - Institute of Product Engineering, Karlsruhe Institute of Technology (KIT), Germany. \email{nehal.afifi@kit.edu}}

\author{Abdelmonem Elhendawi}
\address{SUPMICROTECH-ENSMM, France. \email{ali.abdelmonem@ens2m.org}}

\author{Felix Leitenberger}
\address{IPEK - Institute of Product Engineering, Karlsruhe Institute of Technology (KIT), Germany. \email{felix.leitenberger@kit.edu}}

\author{Nadine Piat}
\address{SUPMICROTECH-ENSMM, France. \email{nadine.piat@ens2m.fr}}

\author{Sven Matthiesen}
\address{IPEK - Institute of Product Engineering, Karlsruhe Institute of Technology (KIT), Germany. \email{sven.matthiesen@kit.edu}}

\begin{abstract}
Electro-Hydrostatic Actuators (EHAs) are widely used in aerospace and industrial systems, where timely detection of sensor anomalies is essential to ensure safe and reliable operation. However, the large volume and high sampling frequency of EHA sensor data pose challenges for accurate and efficient anomaly detection. Conventional statistical and classical machine-learning methods such as Z-score, Interquartile Range (IQR), Median Absolute Deviation (MAD), Isolation Forest, Gaussian Mixture, and k-means often fail to capture the temporal dependencies inherent in EHA signals, resulting in limited detection accuracy and elevated false-alarm rates. Furthermore, systematic evaluations of data-driven anomaly detection approaches for EHA systems remain scarce, particularly under varying operational conditions.This study presents an offline anomaly-detection framework for univariate EHA sensor signals, focusing on temperature and pressure data collected from a controlled test bench. The method employs a reconstruction-based Long Short-Term Memory (LSTM) autoencoder, calibrated and evaluated using validation-set reconstruction-error distributions.Performance is assessed across multiple fault-injection scenarios using accuracy, precision, recall, and F1-score, complemented by sensitivity analyses under varying operating conditions. The LSTM autoencoder achieved an average accuracy of 99.0\%, precision up to 100\%, recall between 90.2\% and 99.6\%, and F1-scores from 93.1\% to 99.8\%, demonstrating high detection sensitivity and a very low false-alarm rate across all evaluated sensors.These results highlight the feasibility of data-driven offline anomaly detection for EHAs. Future work will focus on adapting the developed framework for an online (real-time) environment, enabling the algorithm to detect anomalies while the system is operating. The goal is to achieve near-real-time performance with minimal latency and maintain high accuracy under varying operational conditions.
\end{abstract}
]
\section{Introduction}
Electro-Hydrostatic Actuators (EHAs) are increasingly deployed in aerospace and industrial systems due to high power density, compactness, and improved energy efficiency compared to conventional hydraulic actuation solutions ~\cite{EHAA}. Because EHAs often operate in safety-critical contexts, reliable monitoring is essential: sensor anomalies can corrupt state estimation, degrade performance, or mask developing faults. Modern EHA test benches and fielded systems are instrumented with multiple sensors at high sampling rates, producing large volumes of time-series measurements during operation ~\cite{act11110308}, which makes robust anomaly detection both important and challenging. Classical statistical methods (e.g., Z-score, IQR, MAD) and traditional machine-learning approaches (e.g., clustering, mixture models, Isolation Forest) typically rely on stationarity or weak temporal assumptions and therefore struggle to represent the temporal dependencies and nonlinear dynamics characteristic of EHA signals ~\cite{khan2025enhancing}. Reconstruction-based deep learning, and in particular LSTM-based autoencoders, offers a practical alternative because LSTM architectures are designed to capture long-range temporal dependencies in sequential data ~\cite{lls,ZHANG2021144507}. However, systematic studies applying LSTM-based anomaly detection to EHA systems, especially under controlled operating conditions and reliability-oriented evaluation, remain limited. This paper addresses this gap by proposing an offline LSTM autoencoder framework for univariate EHA sensor signals, calibrated using validation reconstruction-error distributions and evaluated under fault-injection scenarios across operating conditions.
\section{Related Work}
Anomaly detection for sensor and actuator systems is widely studied for condition monitoring in safety-critical domains. Because labelled fault data are scarce in operational settings, most practical methods adopt unsupervised or semi-supervised formulations. Early approaches rely on statistical and shallow-learning techniques such as threshold monitoring, PCA variants, density estimation, and clustering \cite{Paffenroth2018}. While efficient and interpretable, these techniques often assume stationarity or linear structure and can degrade under nonlinear and temporally correlated actuator telemetry.

Classical machine-learning extends this line via one-class learners and forecasting-based detection. One-Class SVM and Support Vector Data Description model nominal behaviour using a learned boundary \cite{Tax2004}, while autoregressive forecasting flags anomalies through large prediction residuals \cite{Guennemann2014}. However, these methods typically struggle to scale to high-frequency time-series and to represent long-horizon temporal dependencies.

Deep-learning-based approaches address these limitations by learning temporal representations directly from data. Reconstruction-based autoencoders detect anomalies using elevated reconstruction error \cite{Chen2018}, and recurrent encoder--decoder architectures (notably LSTM-based) further improve performance by explicitly modelling sequence dynamics \cite{Zhang2021,Filonov2017}. Variants using convolutional components, variational objectives, or adversarial training have shown strong results across multiple domains \cite{Memarzadeh2020,Siegel2020}, while hybrid frameworks combine reconstruction, forecasting, and density estimation to improve robustness \cite{Zong2018}. A representative LSTM-autoencoder thresholding pipeline demonstrated strong accuracy and robustness on real-world environmental sensor data \cite{10011213}, motivating similar temporal architectures for actuator monitoring.

Across domains, unsupervised detectors commonly assume abnormal behaviour is statistically distinct from nominal behaviour \cite{932195}. This motivates clustering-based anomaly identification \cite{4085803} and probabilistic approaches based on low likelihood under Bayesian models \cite{10.1007/978-3-540-77690-1_6}, although distributional drift remains a persistent limitation. To mitigate drift and deployment constraints, hybrid frameworks fuse statistical residual analysis with deep forecasting or reconstruction signals \cite{s19112451}. Survey studies summarize trade-offs among accuracy, computational complexity, and deployability \cite{ERHAN202164}, and recent work emphasizes scalable and efficient pipelines for resource-constrained and industrial IoT settings \cite{9410461,9915308}. Despite these advances, evaluations on actuator-specific telemetry under controlled fault-injection conditions remain limited, particularly for electro-hydrostatic actuators where prior work is often model-based or supervised.

Motivated by these gaps, this work investigates a reconstruction-based LSTM autoencoder for offline univariate anomaly detection in EHA sensor data for actuator health monitoring.
\section{Problem Formulation}
The anomaly detection literature offers mature families of methods for time-series monitoring, including statistical detectors, one-class learning, and deep temporal models. Nevertheless, evidence is still fragmented with respect to actuator health monitoring: many studies emphasize generic time-series settings, while comparatively fewer works report rigorous evaluations on actuator sensor streams acquired under controlled fault conditions and across changing operating regimes. In parallel, labelled fault data remain limited in practice, which constrains purely supervised formulations and motivates approaches that learn nominal behaviour and detect deviations with minimal annotation.

\section{Contribution.} To address this gap, the present work investigates an offline, data-driven anomaly detection framework for \emph{univariate} EHA sensor channels, focusing on capturing the temporal characteristics of nominal operation and identifying anomalous deviations in unseen recordings. The study is supported by a reliability-oriented experimental evaluation on high-resolution test-bench measurements spanning fault-injection scenarios and operating conditions, and includes benchmarking against representative baseline detectors to clarify the practical benefits and limitations of the proposed approach.
\section{Methodology}
\begin{figure*}[t]
    \centering
    \includegraphics[width=0.95\textwidth, height=0.5\columnwidth]{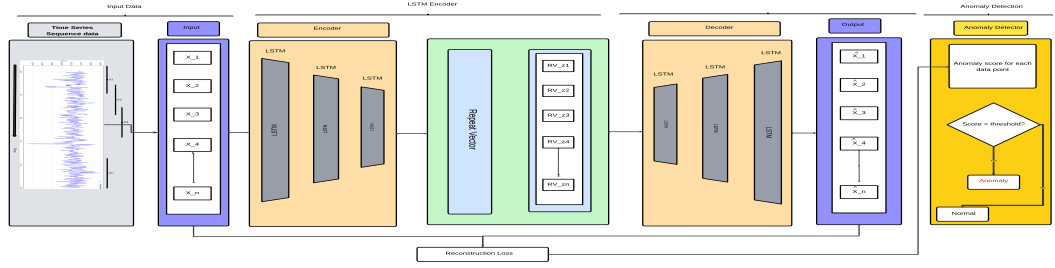}
    \caption{LSTM autoencoder architecture with reconstruction for nominal temporal behaviour.}
    \label{fig:lstm_ae_architecture}
\end{figure*}

We employ an offline, reconstruction-based anomaly detection approach for univariate sensor channels. The underlying assumption is that a model trained on nominal operation reconstructs normal temporal patterns accurately, whereas anomalous deviations lead to increased reconstruction discrepancy. To capture temporal dependencies in the signal, we use an LSTM encoder--decoder (autoencoder) architecture; the model structure is illustrated in Fig.~\ref{fig:lstm_ae_architecture}. Anomalies are detected by scoring each window using reconstruction error and applying a threshold calibrated from nominal data.

\paragraph{Temporal Data Preprocessing.} Let $x_{1:T}=\{x_1,\ldots,x_T\}$ with $x_t\in\mathbb{R}$ denote a univariate sensor signal. The signal is normalized using statistics computed from nominal training data. It is then segmented into overlapping windows of fixed length $L$ with stride $r$. Each window is denoted by
\begin{equation}
\begin{aligned}
\mathbf{x}_i \;&=\; [x_i,\ldots,x_{i+L-1}]^\top \in \mathbb{R}^{L}, \\
i \;&=\; 1,\, 1+r,\, 1+2r,\, \ldots
\end{aligned}
\label{eq:window}
\end{equation}

\paragraph{LSTM Autoencoder and Reconstruction Error.}
Given a window $\mathbf{x}_i$, the encoder maps the sequence to a latent representation and the decoder produces a reconstruction $\hat{\mathbf{x}}_i$. The model is trained on nominal windows only by minimizing the average reconstruction loss
\begin{equation}
\begin{aligned}
(\theta^\star,\phi^\star) \;&=\; \arg\min_{\theta,\phi}\ \frac{1}{N}\sum_{i=1}^{N}\ell(\mathbf{x}_i,\hat{\mathbf{x}}_i), \\
\hat{\mathbf{x}}_i \;&=\; g_{\phi}\!\left(f_{\theta}(\mathbf{x}_i)\right).
\end{aligned}
\label{eq:objective}
\end{equation}
In this work, $\ell(\cdot,\cdot)$ is defined as the mean absolute error (MAE),
\begin{equation}
\begin{aligned}
\ell(\mathbf{x}_i,\hat{\mathbf{x}}_i) \;&=\; \frac{1}{L}\sum_{j=1}^{L}\left|x_{i+j-1}-\hat{x}_{i+j-1}\right|,
\end{aligned}
\label{eq:mae}
\end{equation}
We adopt MAE because it preserves the physical units of the sensor signal and is less sensitive to occasional large spikes compared to squared-error losses, which improved empirical stability in our experiments. Training uses gradient-based optimization with early stopping based on nominal validation loss.

\paragraph{Anomaly Score and Thresholding.}
For each window, we define an anomaly score as its reconstruction error:
\begin{equation}
\begin{aligned}
E_i \;&=\; \ell(\mathbf{x}_i,\hat{\mathbf{x}}_i), \hat{y}_i \;&=\; \mathbb{I}[E_i>\tau].
\end{aligned}
\label{eq:score_decision}
\end{equation}
A decision threshold $\tau$ is calibrated from nominal reconstruction errors (e.g., using an upper-tail cutoff such as a high quantile or a conservative maximum-error rule). Due to window overlap, sustained deviations typically yield consecutive anomalous windows, facilitating event-level localization in offline analysis.

\begin{algorithm}[t]
\label{alg:lstm_ae}
\caption{Offline reconstruction anomaly detection using an LSTM AE}
\begin{algorithmic}[1]
\Require Nominal signal $x_{1:T}$, test signal $x'_{1:U}$, window length $L$, stride $r$
\Ensure Test-window scores $\{E'_i\}$ and labels $\{\hat{y}_i\}$
\State Normalize using nominal training statistics; extract windows $\{\mathbf{x}_i\},\{\mathbf{x}'_i\}$ via (\ref{eq:window}).
\State Train LSTM-AE on $\{\mathbf{x}_i\}$ by minimizing MAE loss (\ref{eq:objective})--(\ref{eq:mae}).
\State Compute nominal errors $E_i=\ell(\mathbf{x}_i,\hat{\mathbf{x}}_i)$ and set threshold $\tau$ from their upper tail.
\ForAll{test windows $\mathbf{x}'_i$}
    \State Reconstruct $\hat{\mathbf{x}}'_i$; compute $E'_i=\ell(\mathbf{x}'_i,\hat{\mathbf{x}}'_i)$.
    \State Predict $\hat{y}_i\gets \mathbb{I}[E'_i>\tau]$ using (\ref{eq:score_decision}).
\EndFor
\end{algorithmic}
\end{algorithm}
\section{Experimental Setup}
All measurements used in this study were acquired on a controlled EHA test bench and measurement infrastructure described in \cite{Leitenberger2021,Dorr2022}. The experimental platform provides repeatable operation under well defined regimes and synchronized high frequency sensing, supporting a reliability oriented evaluation.

\subsection{EHA Test Bench}
\begin{figure}[t]
    \centering
    \includegraphics[width=0.95\linewidth]{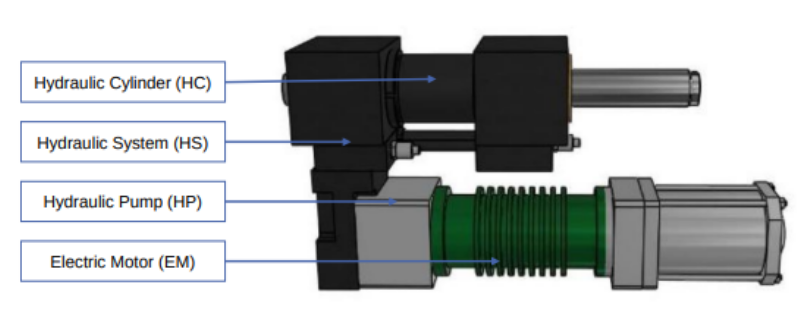}
    \caption{Design of the EHA used in the experimental setup \cite{Leitenberger2021}.}
    \label{fig:eha_design}
\end{figure}

The employed actuator design is shown in Fig.~\ref{fig:eha_design}. The bench is operated using an ADwin Pro2 measurement and control system interfaced via EtherCAT to an external inverter. The outer position control loop runs at 2 kHz using a magnetostrictive position senso, while the internal speed and current loops are implemented on the inverter at 4 kHz and 16 kHz, respectively.

Experiments were conducted under two representative load conditions. The first is an inertial load realized by an attached mass of 11.83 kg. The second is a spring load with stiffness 241.381 N/mm. The bench is instrumented with multiple sensors; this study reports results on univariate temperature and pressure channels recorded as continuous time series at high sampling rate, capturing transient dynamics and regime dependent behaviour.

\subsection{Data Acquisition}
Recordings were checked for invalid or missing samples. Temperature and pressure signals were acquired as continuous time series and prepared consistently with the preprocessing described in the Methodology section. The dataset was split into training and test subsets using a 70/30 ratio. The anomaly detection model was trained exclusively on nominal data.

\subsection{Dataset Construction and Labelling}
To obtain a clean nominal training set, a channel wise sigma filtering rule was applied on fault free recordings. For each sensor channel, the mean $\mu$ and standard deviation $\sigma$ were computed from nominal samples, and observations exceeding
\begin{equation}
\begin{aligned}
x(t) \;&>\; \mu + k\sigma,\;\;\;\; k \;\in\; \{3,4\}
\end{aligned}
\label{eq:sigma_filter}
\end{equation}

\begin{figure*}[t]
    \centering
    \begin{subfigure}[t]{0.42\textwidth}
        \centering
        \includegraphics[width=\linewidth]{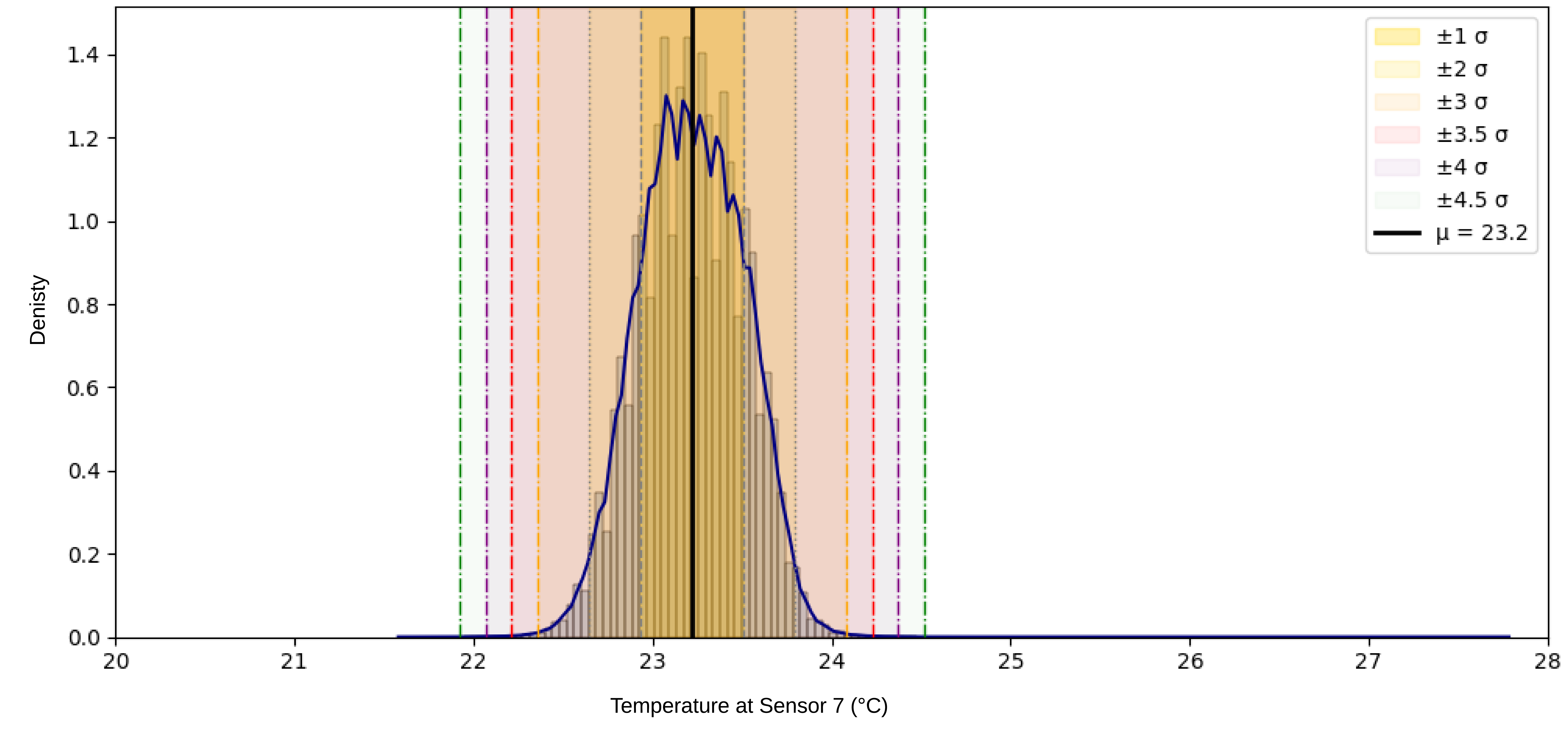}
        \caption{Sensor distribution and $\mu+k\sigma$ bounds.}
        \label{fig:sensor_distribution}
    \end{subfigure}\hfill
    \begin{subfigure}[t]{0.28\textwidth}
        \centering
        \includegraphics[width=\linewidth]{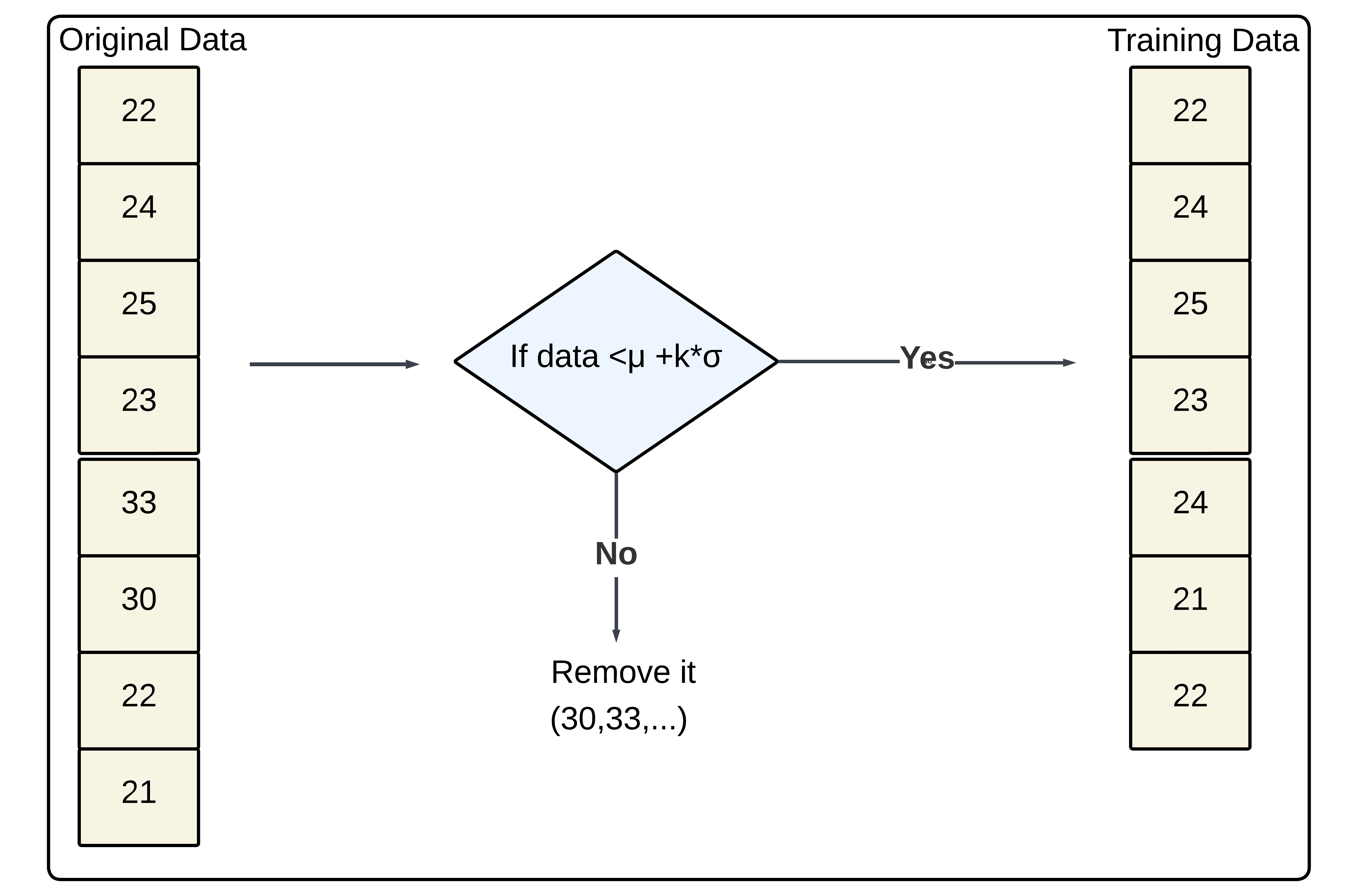}
        \caption{Nominal training set.}
        \label{fig:training_creation}
    \end{subfigure}\hfill
    \begin{subfigure}[t]{0.3\textwidth}
        \centering
        \includegraphics[width=\linewidth]{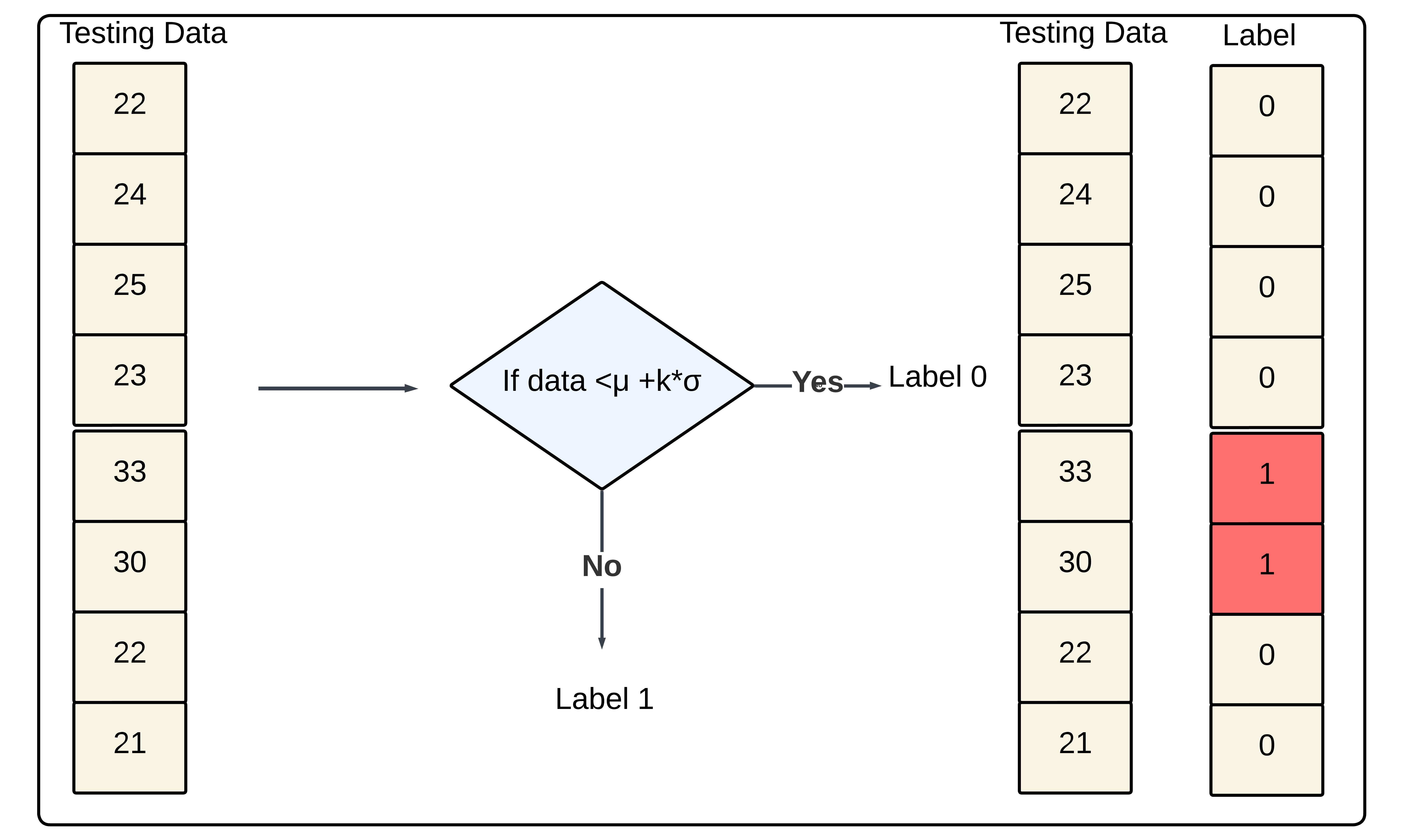}
        \caption{Test set labeling.}
        \label{fig:test_creation}
    \end{subfigure}
    \caption{Dataset preparation and evaluation protocol.}
    \label{fig:dataset_overview}
\end{figure*}

were removed, where $k$ was selected according to channel variability. The test set preserves the full sensor range and contains nominal and anomalous segments. For quantitative reporting, binary evaluation labels follow the same operational definition, with $x(t)\le \mu+k\sigma$ treated as nominal and $x(t)>\mu+k\sigma$ treated as anomalous. These labels are used only for evaluation and do not enter training. Figure~\ref{fig:dataset_overview} summarizes the nominal distribution used to select $\sigma$ bounds, the nominal filtering procedure used for training set construction, and the test set labeling workflow.
\section{Results}
\begin{figure}[t]
    \centering
    \includegraphics[width=0.85\columnwidth]{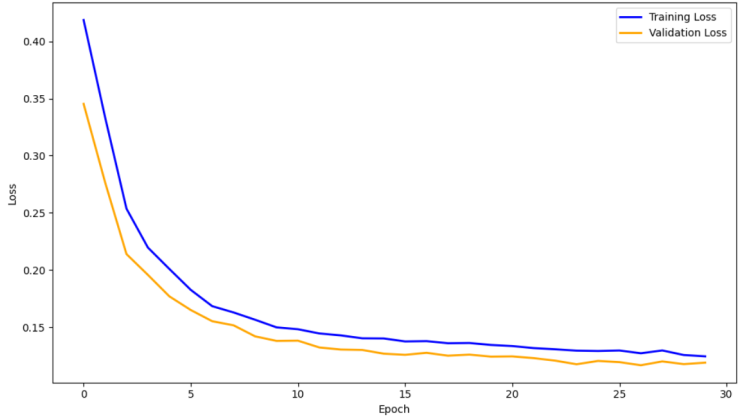}
    \caption{Training and validation loss of LSTM-AE.}
    \label{fig:training_loss}
\end{figure}

\begin{figure}
    \centering
    \includegraphics[width=\columnwidth]{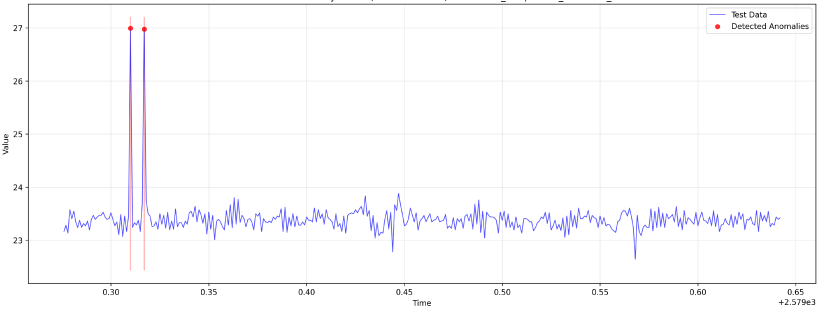}
    \caption{Spike event example: raw sensor signal (blue) and detected anomalies (red).}
    \label{fig:spike_event_1}
 \end{figure}

\begin{figure}
    \centering
    \includegraphics[width=\columnwidth]{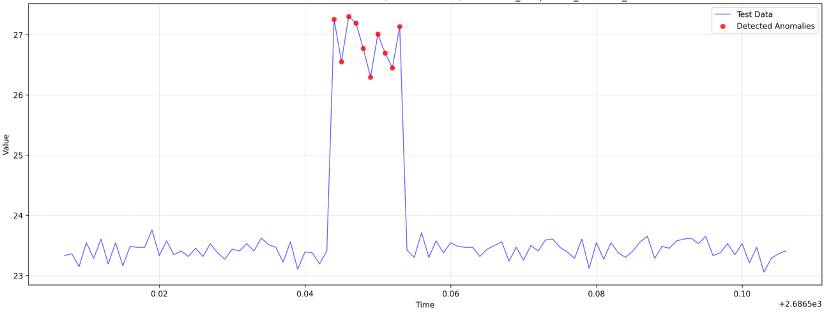}
    \caption{Sustained spike event: raw sensor signal (blue) and detected anomalies (red).}
    \label{fig:spike_event_2}
\end{figure}

The proposed LSTM-AE is evaluated using offline EHA datasets under nominal and fault-injection conditions. The model is trained exclusively on nominal data, with a chronological validation split used for early stopping and threshold calibration. The final training configuration and hyperparameters are summarized in Table~\ref{tab:lstm_training_params}.
\begin{table}[t]
\centering
\caption{LSTM-AE training parameters.}
\label{tab:lstm_training_params}
\footnotesize
\setlength{\tabcolsep}{4pt}
\renewcommand{\arraystretch}{1.15}
\begin{tabular}{p{0.49\columnwidth} c p{0.49\columnwidth}}
\hline
\textbf{Hyperparameter} & \textbf{Value} \\
\hline
Learning rate    & \textbf{0.0001} \\
Dropout          & \textbf{0.20}  \\
Batch size       & \textbf{64}  \\
Epochs (max)     & \textbf{30}  \\
Loss function    & \textbf{MAE} \\
Validation split & \textbf{Last 10\%} \\
\hline
\end{tabular}
\end{table}
Model convergence is illustrated in Fig.~\ref{fig:training_loss}, which shows smooth and stable training and validation loss curves over the training epochs, indicating effective optimisation and good generalisation without overfitting.
Detection performance is assessed using classification metrics, including accuracy, precision, recall, F1-score, and ROC--AUC. To provide a system-level evaluation, results are reported as averages across representative pressure and temperature sensor channels.
\begin{table*}[t]
\centering
\caption{Comparative performance of baseline models and the proposed LSTM-AE, averaged across key EHA sensor signals.}
\label{tab:comparison_baselines}
\setlength{\tabcolsep}{7pt} 
\renewcommand{\arraystretch}{0.85} 

\begin{tabular}{lccccc}
\hline
\textbf{Model} & \textbf{Accuracy} & \textbf{Precision} & \textbf{Recall} & \textbf{F1-score} & \textbf{ROC--AUC} \\
\hline
Z-score               & 0.918  & 0.001 & 0.316 & 0.003 & 0.617 \\
IQR                   & 0.873  & 0.001 & 0.376 & 0.002 & 0.625 \\
MAD                   & 0.979  & 0.003 & 0.175 & 0.006 & 0.577 \\
Isolation Forest      & 0.360  & 0.000 & 0.697 & 0.001 & 0.528 \\
Gaussian Mixture      & 0.905  & 0.001 & 0.341 & 0.003 & 0.623 \\
\textit{k}-means      & 0.905  & 0.001 & 0.343 & 0.003 & 0.624 \\
Local Outlier Factor  & 0.999  & 0.016 & 0.008 & 0.010 & 0.504 \\
AE                    & 0.9997 & 1.000 & 0.249 & 0.398 & 0.624 \\
CNN--AE               & 0.995  & 0.030 & 0.115 & 0.048 & 0.556 \\
\textbf{LSTM--AE}     & \textbf{0.995--0.999} & \textbf{0.96--1.00} & \textbf{0.90--0.99} & \textbf{0.93--0.99} & \textbf{0.95--0.99} \\
\hline
\end{tabular}
\end{table*}
Table~\ref{tab:comparison_baselines} compares the proposed LSTM-AE with statistical, classical machine-learning, and deep autoencoder baselines. Statistical methods and classical models achieve high nominal accuracy but exhibit near-zero precision and F1-scores due to class imbalance and the absence of temporal modelling. Deep autoencoder baselines improve reconstruction fidelity but remain limited in recall and robustness.
In contrast, the LSTM-AE consistently achieves high precision, recall exceeding 90\%, and ROC--AUC values close to unity across sensors, demonstrating a substantially improved balance between sensitivity and false-alarm suppression.
Qualitative inspection supports the quantitative findings. Fig.~\ref{fig:spike_event_1} and Fig.~\ref{fig:spike_event_2} show representative spike events, where abrupt transients in the raw sensor signal are correctly identified as anomalies. Detected anomalies form coherent segments rather than isolated points, while normal fluctuations before and after the event are not falsely flagged. This highlights the benefit of temporal context in suppressing spurious alarms.
\begin{table*}[t]
\centering
\caption{Comparison with representative anomaly detection approaches reported in the literature.}
\label{tab:literature_comparison}
\small
\setlength{\tabcolsep}{4pt}
\renewcommand{\arraystretch}{0.9}
\begin{tabular}{p{0.22\textwidth} p{0.18\textwidth} p{0.24\textwidth} c c c c}
\hline
\textbf{Reference} & \textbf{Technique} & \textbf{Dataset} & \textbf{Acc.} & \textbf{Prec.} & \textbf{Rec.} & \textbf{F1} \\
\hline
~\cite{Yin2022}        & LSTM--AE              & Yahoo Webscope S5      & 99.25 & 97.84 & 94.16 & 95.97 \\
~\cite{Nguyen2021} & LSTM--AE + OCSVM      & Generated dataset      & 98.36 & 98.45 & 95.99 & 96.98 \\
~\cite{Li2019}         & Adversarial LSTM--AE  & CCV (video)            & 83.03 & 75.08 & 73.26 & 74.16 \\
~\cite{Sharma2020} & LSTM--AE              & CERT insider threat    & 90.17 & --    & 91.03 & --    \\
~\cite{Wei2023}       & LSTM--AE              & CO$_2$ dataset          & 99.50 & 100  & 89.90 & 94.68 \\
~\cite{Tran2020}     & LSTM--AE + iForest    & Fashion data           & 95.00 & 100  & 94.00 & 87.00 \\
\textbf{This work}              & \textbf{LSTM--AE}     & \textbf{EHA sensor data} & \textbf{99.0--99.8} & \textbf{99--100} & \textbf{93--98} & \textbf{93--98} \\
\hline
\end{tabular}
\end{table*}

\section{Discussion}
The results indicate that modeling temporal dependencies with an LSTM AE yields more reliable anomaly detection on EHA sensor streams than non temporal baselines. The main contribution is a practical detection pipeline that achieves a favorable precision recall trade off across both pressure and temperature signals, and qualitatively produces contiguous anomalous segments rather than isolated outliers, which is more suitable for condition monitoring. 
The baseline comparison also highlights that high accuracy is not informative under class imbalance, as several detectors appear accurate while exhibiting near zero precision and F1 score. In contrast, the LSTM AE improves precision, recall, and ROC AUC, consistent with improved separability when temporal structure is represented.
Three key limitations were noted, labels are derived from a transparent $\sigma$-rule, yet may not reflect fault criticality or slow degradation that remains within nominal bounds; the analysis is univariate and therefore does not exploit cross-channel dependencies; and performance depends on threshold calibration, which may require retuning under regime shifts or sensor drift. Addressing these in future work would strengthen the link between anomaly detection outputs and maintenance decision-making.
\section{Conclusion}
This work developed an offline anomaly detection pipeline for unlabeled EHA sensor time-series using a univariate LSTM autoencoder. The model learns nominal temporal dynamics from data and flags deviations through reconstruction error, avoiding the need for fault labels during training. The workflow includes sigma-based cleaning, z-score normalization, sliding-window segmentation, and validation-based threshold selection to support reproducible evaluation. Test-bench experiments with expert-defined fault injections showed high detection performance across operating conditions, with accuracy above $99\%$ and low false-alarm rates relative to statistical and conventional machine-learning baselines, indicating that reconstruction-based sequence models are suitable for offline EHA condition monitoring. Key limitations are that evaluation labels follow an operational $\sigma$-rule that may not reflect fault criticality or slow within-bounds degradation, the univariate setup does not capture cross-channel dependencies, and thresholds may require retuning under regime shifts or sensor drift. Future work will address these points by extending the method to real-time deployment (e.g., compression and low-latency inference), integrating multivariate sensor fusion, broadening fault coverage for validation, and assessing robustness under long-term operational conditions.
\renewcommand\theequation{A.\arabic{equation}}
\begin{acknowledgement}
This research was funded by the Federal Ministry for economic Affairs and Climate Action
under the funding number 20Y1910E. While preparing this work, the authors used AI tools such as deepl.com and Gemini 3 to improve readability and language. After utilizing these tools, the authors reviewed and edited the content as necessary. The authors take full responsibility for the publication’s content.
\end{acknowledgement}
\bibliographystyle{chicago}
\bibliography{References}

\end{document}